\documentclass[letterpaper, 10pt, conference]{ieeeconf}

\usepackage{amsmath,amssymb,amsfonts}
\usepackage{graphicx}
\usepackage{booktabs}
\usepackage{hyperref}
\usepackage{subcaption}
\newcommand{\num}[1]{#1}

\title{Risk-Aware Preference Learning for Stochastic Outcomes}

\author{Yi-Shiuan Tung, Yuni Wu, Wei Jiang, Alessandro Roncone, Bradley Hayes\\
Department of Computer Science, University of Colorado Boulder\\ \{\tt\small yi-shiuan.tung, yuni.wu, wei.jiang, alessandro.roncone, bradley.hayes\}\\@colorado.edu}

\begin{document}
\maketitle

\begin{abstract}
Learning reward functions from human preferences is a widely used approach for aligning robot behavior with user expectations in human-robot interaction. Most existing approaches assume that humans evaluate uncertain outcomes using expected utility (EU), aggregating outcome utilities linearly with their probabilities. However, behavioral evidence shows that humans are systematically risk-sensitive, overweighting rare negative events and exhibiting loss aversion. We study the consequences of this mismatch in social robot navigation, where safety-critical outcomes (e.g., collisions) are rare but highly consequential. We compare EU with Cumulative Prospect Theory (CPT), a nonlinear model of human decision-making, within a Bradley-Terry preference learning framework. Our preliminary experiments show that when preferences are generated by risk-sensitive users, CPT-based learners recover reward functions with substantially lower regret compared to EU-based learners. Our results highlight the importance of modeling human risk sensitivity when learning rewards from preferences over stochastic robot outcomes.
\end{abstract}

\section{Introduction}

Preference-based reward learning has emerged as a powerful paradigm for aligning autonomous systems with human intent~\cite{christiano2017deep,sadigh2017active}. In domains such as social robot navigation, robots must balance efficiency, safety, and adherence to social norms, objectives that are difficult to specify manually but natural for humans to express through comparisons \cite{tung2026CRED}. Most existing approaches assume that humans evaluate uncertain outcomes using \emph{expected utility} (EU), where the value of an action is the probability-weighted sum of outcome utilities. However, decades of behavioral economics research show that humans systematically deviate from EU~\cite{kahneman1979prospect,tversky1992advances}. In particular, humans overweight rare catastrophic events and exhibit loss aversion. These biases are particularly relevant in social navigation, where safety-critical outcomes like collisions are rare but consequential, and where human evaluators are known to be risk-sensitive~\cite{kruse2013human, mavrogiannis2023core}.

\begin{figure}[t]
    \centering
    \begin{subfigure}[b]{0.45\linewidth}
        \includegraphics[width=\linewidth]{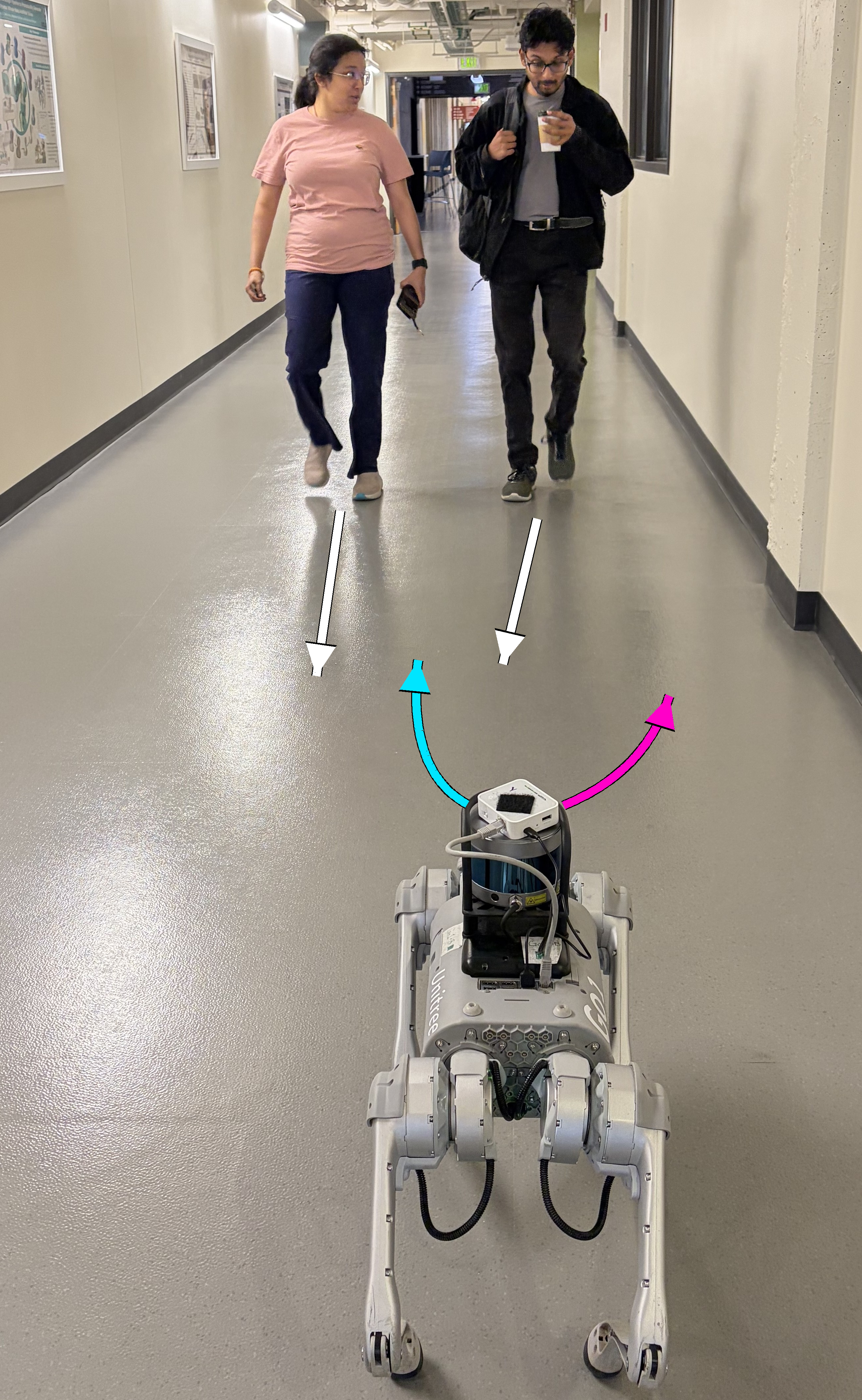}
    \end{subfigure}
        \begin{subfigure}[b]{0.45\linewidth}
        \includegraphics[width=\linewidth]{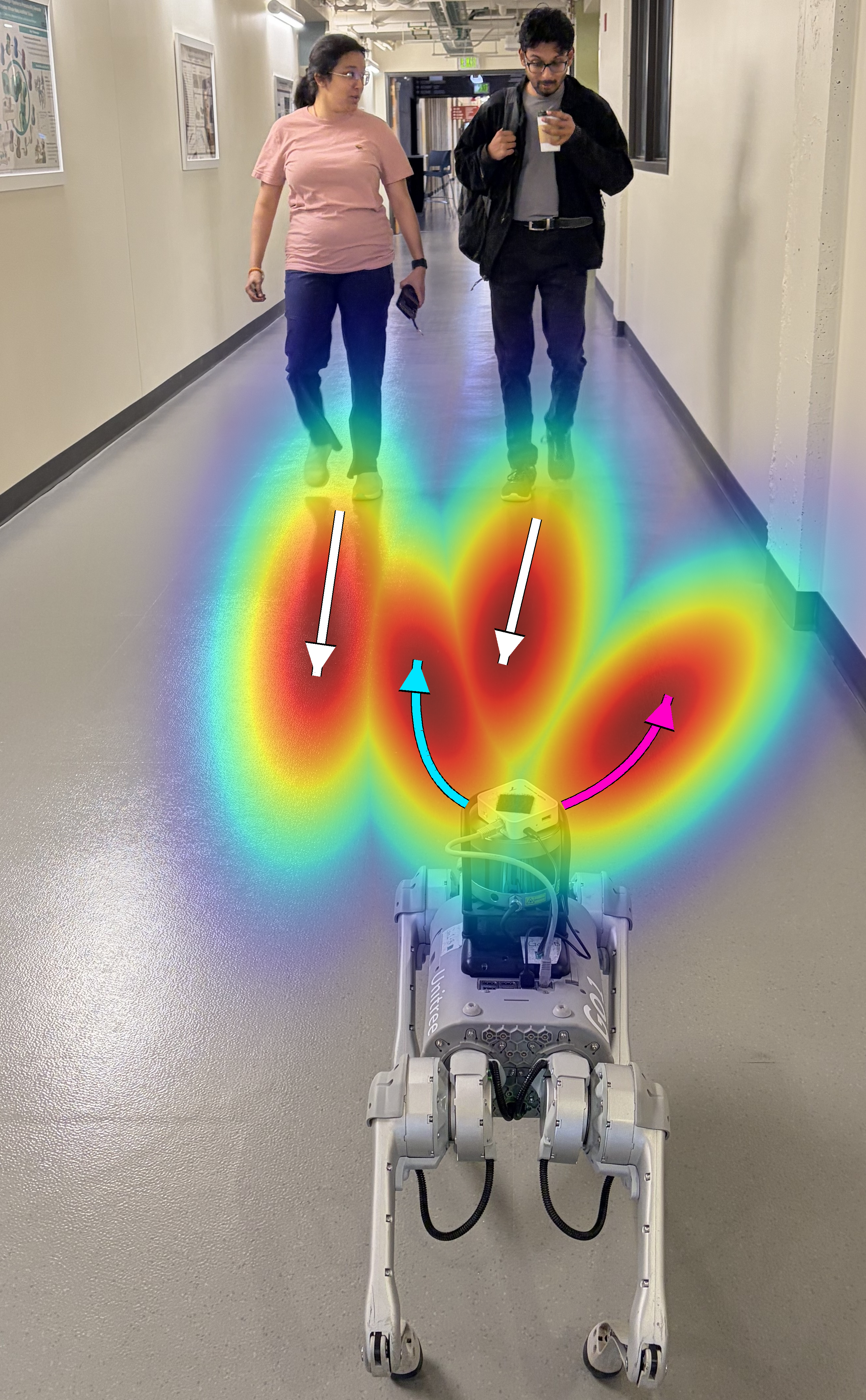}
    \end{subfigure}
    \label{fig:intro}
    \caption{Standard preference queries often show deterministic trajectories (left), while real world navigation induces distributions over stochastic outcomes (right). We study whether modeling user risk sensitivity with CPT improves reward learning from preferences over such stochastic outcomes.}
\end{figure}

We hypothesize that this mismatch leads to structural errors in reward learning.
When the true human decision-maker is risk-sensitive but the learner assumes EU, the learned reward function may incorrectly encode risk sensitivity as differences in outcome utility, yielding a biased reward that conflates what users value with how they weight uncertainty. We propose modeling human preferences over stochastic robot outcomes using \emph{Cumulative Prospect Theory} (CPT), which captures nonlinear probability weighting and asymmetric valuation of gains and losses. We compare this model against a standard EU-based preference learner in a 2D social navigation task. Our results show that, when the user is risk-sensitive, the CPT learner recovers a reward function with lower regret than the EU learner. These findings suggest that modeling risk sensitivity can improve reward learning under uncertainty; human-subject validation remains future work.

\section{Related Work}
Learning reward functions from human preferences is a common approach for aligning robot and machine-learning systems with human intent~\cite{christiano2017deep,sadigh2017active,biyik2018batch}. These methods assume expected utility, which can conflate what users value with how they weight uncertainty in stochastic domains.

Risk-sensitive decision-making has been studied through objectives such as Conditional Value-at-Risk (CVaR)~\cite{tamar2015policy,chow2017risk}. These methods make the robot's planner risk-aware, but they do not model risk sensitivity in the human preference-generation process. Cumulative Prospect Theory (CPT) offers a descriptive model of human decision-making under uncertainty ~\cite{tversky1992advances,prelec1998probability}. While CPT has been used to model behavior in driving and interaction settings~\cite{sun2019interpretable}, its role in learning reward functions from human preferences remains underexplored. We address this gap by comparing EU- and CPT-based preference learners for stochastic robot outcomes.
\section{Approach}
We consider the problem of learning a reward function from pairwise human preferences over robot behaviors. Let $\phi(o) \in \mathbb{R}^d$ denote a feature vector describing outcome $o$, and let $\theta \in \mathbb{R}^d$ be the reward weight vector to be learned. Each robot action induces a distribution over outcomes: action $A$ produces outcome $o_i$ with probability $p_i$, where $\sum_i p_i = 1$.
Given two actions $A$ and $B$, we model the probability that a human prefers $A$ over $B$ using the Bradley-Terry model~\cite{bradley1952rank}: 
\begin{equation}
P(A \succ B) = \sigma\!\big(\beta (V(A) - V(B))\big),
\end{equation}
where $\sigma(\cdot)$ is the sigmoid function, $\beta > 0$ is an inverse temperature controlling the noise level in preferences, and $V(\cdot)$ is the value function that maps an action to a scalar.

Under EU, the value of an action is a linear expectation over outcome utilities: $V_{\text{EU}}(A) = \sum_i p_i \, \theta^\top \phi(o_i)$. The Cumulative Prospect Theory (CPT) replaces this linear aggregation with nonlinear transformations. First, utilities are transformed using a value function:
\begin{equation}
v(x) =
\begin{cases}
x^\alpha & x \geq 0, \\
-\lambda (-x)^\eta & x < 0,
\end{cases}
\end{equation}
where $\lambda > 1$ captures loss aversion. Second, probabilities are distorted using a weighting function: $w(p) = \exp\!\big(-(-\ln p)^\gamma\big)$,
which overweights rare events when $\gamma < 1$~\cite{prelec1998probability}. Decision weights $\pi_i$ are computed via a rank-dependent transformation over cumulative probabilities~\cite{tversky1992advances}.
The CPT value of an action is: $V_{\text{CPT}}(A) = \sum_i \pi_i \, v\big(\theta^\top \phi(o_i) - r\big)$, where $r$ is a reference point that divides gains and losses.
\section{Experiments}
We evaluate whether explicitly modeling risk-sensitive preferences improves reward recovery in stochastic social navigation. We construct a 2D simulation environment in which a robot selects navigation actions while pedestrian behaviors are generated by the Helbing--Moln{'a}r social force model \cite{helbing1995social}, implemented with \texttt{pysocialforce}. Our data consists of navigation scenes spanning corridors, intersections, doorways, and open spaces with diverse interaction patterns such as head-on encounters and group blocking.
\begin{table}[t]
\centering
\caption{CPT parameters for synthetic users.}
\label{tab:cpt_params}
\begin{tabular}{lccccc}
\toprule
Risk profile & $\alpha$ & $\eta$ & $\lambda$ & $\gamma$ & $\delta$ \\
\midrule
T\&K     & 0.88 & 0.88 & 2.25 & 0.61 & 0.69 \\
Strong & 0.80 & 0.80 & 3.00 & 0.50 & 0.50 \\
\bottomrule
\end{tabular}
\end{table}
\textbf{Setup.}
We model robot actions in dynamic human environments, where each action induces a distribution over outcomes. A robot \emph{action} is one of seven meta-actions (\textit{forward}, \textit{slow\_down}, \textit{stop}, \textit{turn\_left}, \textit{turn\_right}, \textit{forward\_left}, \textit{forward\_right}), each parameterized by velocity $v$ and angular velocity $\omega$. For each scene and robot action, we simulate $K=50$ stochastic rollouts, where each rollout produces a feature vector $f \in \mathbb{R}^5$ summarizing minimum pedestrian clearance, pedestrian deviation induced by the robot, robot progress towards the goal, path smoothness, and collision count. The ground-truth reward is $\theta^\star = [0.10, -0.20, 0.30, 0.05, -0.25]$ over the five features. The empirical distribution over the $K$ rollouts is the action's outcome distribution $\{(f_k, p_k)\}$.

\textbf{Reward Learning.}
We evaluate preference learning under different assumptions about the preference-generating process. The synthetic user or teacher, generates pairwise comparisons using either an EU model with linear utility or a CPT model with nonlinear value and probability weighting. For CPT teachers, we consider two risk-sensitivity profiles: T\&K, based on the empirical parameter values reported by Tversky and Kahneman ~\cite{tversky1992advances}, and Strong, which amplifies risk sensitivity. Table~\ref{tab:cpt_params} summarizes the CPT parameter values.
We then train two classes of learners from the resulting comparisons: an EU learner, which assumes $V=\theta^\top \phi$, and a CPT learner, which jointly estimates the reward weights $\theta$ and CPT parameters. Both learners are trained by minimizing the Bradley--Terry negative log-likelihood. This setup allows us to test whether a learner with the correct preference model recovers lower-regret rewards when the synthetic user exhibits EU or CPT risk sensitivity.

\begin{figure}[t]
\centering
\includegraphics[width=\columnwidth]{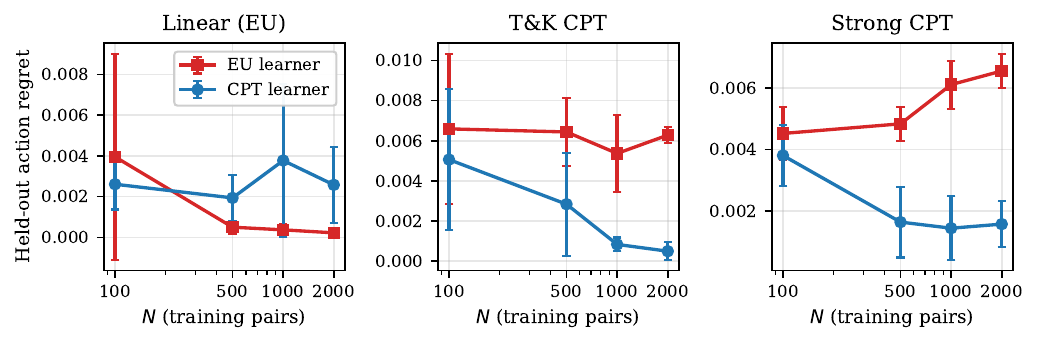}
\caption{Held-out action regret versus training set size $N$ for EU and CPT synthetic teachers (mean $\pm$ std over \num{5} seeds). The EU learner achieves lower regret when preferences are generated by an EU teacher (left), while the CPT learner achieves lower regret for CPT teachers, suggesting that matching the learner's preference model to the teacher's risk-sensitive choice process improves reward recovery.}
\label{fig:learning-curves}
\end{figure}

\section{Results \& Discussion}

Performance is measured by \emph{action regret}, defined as the difference between the value of the action chosen by the true user model and the value of the action selected by the learned model, evaluated on held-out scenes. Fig.~\ref{fig:learning-curves} shows action regret as the number of training preference pairs increases. When preferences are generated by the EU teacher, the EU learner achieves the lowest regret. This is expected: when the teacher is well described by expected utility, the additional CPT parameters introduce unnecessary flexibility and can make learning less sample-efficient.

In contrast, when preferences are generated by CPT teachers, the CPT learner substantially outperforms the EU learner. For the T\&K and Strong CPT teacher, the CPT learner's regret decreases steadily with more preference data, while the EU learner remains at a higher regret level. These results suggest that when users make risk-sensitive choices, an EU learner may explain those choices by distorting the recovered reward function, while a CPT learner can separate reward weights from risk sensitivity. 

Overall, the learning curves provide preliminary evidence that explicitly modeling CPT-style preference formation improves reward learning under stochastic outcomes, especially when risk sensitivity is strong.

\bibliographystyle{IEEEtran}
\bibliography{main}

\end{document}